# Serial-parallel Multi-Scale Feature Fusion for Anatomy-Oriented Hand Joint Detection

Bin Li, Hong Fu, Ruimin Li and Wendi Wang

*Abstract* — **Accurate hand joints detection from images is a fundamental topic which is essential for many applications in computer vision and human computer interaction. This paper presents a two-stage network for hand joints detection from single unmarked image by using serial-parallel multi-scale feature fusion. In stage I, the hand regions are located by a pre-trained network, and the features of each detected hand region are extracted by a shallow spatial hand features representation module. The extracted hand features are then fed into stage II, which consists of serially connected feature extraction modules with similar structures, called "multi-scale feature fusion" (MSFF). A MSFF contains parallel multi-scale feature extraction branches, which generate initial hand joint heatmaps. The initial heatmaps are then mutually reinforced by the anatomic relationship between hand joints. The experimental results on five hand joints datasets show that the proposed network overperforms the state-of-the-art methods.**

## I. INTRODUCTION

Hand joints detection aims to locate keypoints of all hands in an image, such as wrist, fingertips and knuckles. Accurate hand joints detection based on single unmarked image conduces to efficient and robust hand gesture analysis and hand pose reconstructions. It has become an important research topic in many applications, including human computer interaction [1-2], motion recognition [3], sign language interpretation [4-5]. Recently, many significant solutions for hand joints detection have been proposed [6-7]. However, the accuracy and robustness of the detection algorithms are not satisfactory, due to hand gesture variability, background and illuminance change, and unfixed hand scales, etc..

Existing methods for hand joints detection are either based on single image or multi-source images. Single image based methods only need one color image, and multi-source based methods may make use of depth image, mask template, multi-views or specific markers for hand keypoints detection. In this paper, we focus on the method based on single image. To achieve an accurate and robust hand joint detection algorithm, the following issues should be considered. First, the method should be robust to the number of hands and scale variability. The existing hand joint detection is usually supposed to be conducted on a cropped image region which contains single hand with fixed scale [8]. The detection of the hand regions in unfixed scales are not fully investigated. If an image has more than one hand with different scales caused by the view distance or the physical sizes of the hands, the algorithm should be able to locate the hand region accurately, robustly and simultaneously. Second, the feature maps from different scales should be sophisticatedly utilized to achieve better detection results. Large-scale feature maps contain detailed joints information, and small-scale feature maps include global hand information. A self-adaptive network structure to trade-off different scale feature maps with attention mechanism could make the network focus on important features and therefore be robust in hand joint detection in real scenarios. Third, shallow-level feature maps (generated in the early layers of a neural networks) may help the network locate the visible hand joints quickly and accurately, while the deep-level feature maps may conduce to predict the challenging joints which are hidden or occluded. When the network layers and parameters increase, some of those feature maps may become redundant in the network training processes. A proper loss function to take advantage of those multi-level feature maps is also an issue to be considered.

Regarding the issues above, it could be difficult to locate the joints of hands effectively from a single unmarked image by only using existing CNN models. Therefore, a two-stage network using serial-parallel multi-scale feature fusion is presented, which can accurately and efficiently locate hand joints in single image without additional information (e.g. depth, mask etc.). The proposed network is an end-to-end structure, which contains two stages, as shown in Fig.1. In Stage I, a pre-trained model is employed to detect the hand regions, and then the features of each detected hand region are extracted by a module named "shallow spatial hand features representation (SSHFR)". The extracted hand features are then fed into Stage II, which consists of a series of cascaded feature extraction modules with similar structures, called "multi-scale feature fusion" (MSFF). Each MSFF contains parallel multi-scale feature extraction branches, which generate initial hand joint heatmaps. The initial heatmaps are then mutually reinforced by the anatomic connections between hand joints. The output of

Bin Li is with the School of Information Science and Technology, Northwest University, Xi'an, China, 710127 (e-mail: lib@nwu.edu.cn).
Hong Fu (corresponding author) is with Department of Mathematics and Information Technology, The Education University of Hong Kong, Hong Kong, China (e-mail: hfu@eduhk.hk).
Ruimin Li is with Xi'an Institute of Optics and Precision Mechanics, CAS, Xi'an 710119, China (e-mail: liruimin2015@opt.cn).
Wendi Wang is with Shanghai University of Engineering Science, China. (e-mail: wang_wendi@qq.com).

each MSFF is passed to its successor, until stable results are achieved. A loss-wise function is adopted to balance the weight of each MSFF for accurate hand joint detection. In the experiments on both synthetic and real-world datasets, the proposed network has the advantages of high efficiency, strong robustness and excellent performance.

Comparing to other key point detection datasets like body skeleton, manually annotating hand joints are more complicated due to the variability of hand gesture, self-occlusion, light and camera viewpoint changes, so the number of annotated natural hand joint datasets are very limited. MPII-hand [11] and NZSL [12] are commonly used natural hand image datasets. The former contains 1,700 and the latter contains 1,500 images only. RHD [9] and HS [10] contain large amount of hand images, but they are synthetic and their characteristics are quite different from those of the real-world hand images. Therefore, we developed our own natural hand image dataset DCD8-6000, which will be introduced in detail in *Section IV*.

The main contribution of our work includes:
- A two-stage self-adaptive cascaded multiscale network is proposed, which is able to locate all the hands and detect hand joints from a single unmarked image;
- A serially-parallelly connected MSFF structure is proposed, which aggregates both deep features and multi-scale features;
- A mutual reinforcement mechanism is proposed to enhance the heatmaps, oriented by anatomic relationship between hand joints;
- A new real-world dataset DCD8-6000 with annotated hand joints location is presented, which contains 6,000 images from 8 subjects;
- The experiments are conducted on five hand joint datasets to verify the effectiveness of the proposed network.

The rest of this paper is organized as follows. Section II reviews the hand joints detection algorithms in recent years. In Section III, the proposed network is described in detail. The experimental results and discussions are presented in Section IV. Finally, Section V concludes the paper.

## II. RELATED WORK

### A. *Hand joints detection*

The state-of-the-art human hand joints detection algorithms can be summarized into two main categories: single image based and multi-source based.

Early single image based algorithms mainly used hand-craft features to locate hand keypoints. The texture, color and spatial information in the image were extracted by constructing various of robust feature detectors [13][14], *e.g.*, hand contour template, skin model or spatial feature vector. Based on the above characteristics, some researchers utilized an additional geometric skeleton model of the hand to improve the robustness and accuracy of hand joints detection [15]. Zhou et al. [16] proposed a real-time hand skeleton recognition method that extracted finger skeletons from salient hand edges and geometrical characteristics. A skeleton-based approach for hand gesture recognition was shown by Smedt et al. [17]. They used multi-level representation of Fisher Vectors and geometric features to produce the hand joint feature vectors. Then the final joint locations were achieved by a linear classifier.

Different from human pose estimation which focus on gross motion, the detection of hand joint is actually fine motion localization. The hand images are often affected by variances of scale, self-occlusion and low-resolution. Therefore, it is difficult for the traditional hand-craft feature based methods to achieve accurate and efficient result in practical applications. With the development of machine learning, especially deep learning. Researchers have proposed various fancy hand joint detection methods that are based on convolutional neural network structures [18][19]. The CNN-based model has the advantages of generating hidden features of the object, establishing spatial contact between multi-dimensional information and solving the occlusion robustly [20][21]. A deep convolutional pose detection architecture (CPM) was proposed by [22] which was based on the theory of pose machine [23]. They combined image features with contextual features to detect the hand joint positions. Zimmermann et al. [24] continued to refine the CPM structure. An encoder-decoder network architecture, PoseNet, was designed to generate score map of each hand joint. In order to restore hand joint location and skeleton model from the score maps accurately, Cao et al. [25] presented Part Affinity Fields (PAFs) to learn to associate joint parts with individuals in an image.

The model parameters will become complex when the number of layers of deep convolutional neural network increases, which makes the model slow and hard to converge. In order to optimize the network structure and achieve hand joints detection results accurately and efficiently [26-28], Wang et al. [29] presented a real-time hand joint detection network. They used a lightweight bounding box and a cascaded CNN to locate the hand region and joints positions respectively. Oberweger [30] utilized a recurrent network structure to improve hand joint detection efficiency.

It is difficult to effectively solve the problem of self-occlusion of hand joints through a single camera-view. Therefore, the multi-source based hand joints detection algorithms make use of various input information to build a robust hand joints estimation model. Simon et al. [10] presented an approach that used a multi-camera system to train fine-grained detectors for hand keypoints estimation. This multi-image captured method can detect hand joints from different perspectives. Choi et al. [31] obtained the hand joints features from the object-oriented perspective and the hand-oriented perspective respectively. They simultaneously trained two deep neural networks using paired depth images, and the proposed networks share intermediate observations produced from different perspectives to create a more informative representation. Ge et al. [32] reported to utilize multi-view projections to regress for 2D heat-maps and estimated the hand joint locations on different planes. Sridhar et al. [33] proposed a method that can track the full skeleton motion of the hand from multiple RGB cameras in real-time. Joo et al. [34]

presented a multi-view system for human motion detection. The system composed of a massively multi-view camera system in a modularized design and fused those multiple views for robust skeletal pose estimation. The multi-view hand images improve the accuracy of hand joints detection, but it also causes problems such as the increase of the number of training data, and images from different views need to be calibrated and fused. [35] proposed to use RGB-D camera capture color images and depth images simultaneously for hands joint detection. It can effectively solve the problem of multi-view images matching. The multi-source based methods have achieved good performance, but they actually rely on the detection results from each single image. Therefore, an accurate, robust and efficient hand joints detection method which only uses single image is still a fundamental problem.

B. *Multiscale Feature Fusion structure*

The multiscale feature fusion structure is divided into two categories, which are parallel multi-branch fusion and cascaded residual connection structure. They all aim to extract the feature maps of the target in different perceptive fields.

In order to generate multi-scales feature maps in parallel, researchers build a series of branches with similar structure and use different scale convolution kernels to generate multi-scale feature maps [39, 40]. Some also use the spatial pyramid pooling structure [41, 42], or hole convolution to obtain different scale feature maps [43]. The parallel structure generates feature maps at the same network layer, while the cascaded structure can fuse the features of different layers. The cascaded structure realizes the combination of global and local information, effectively expands the receptive field, and refines the representation ability of the feature map [44-46].

Different from most existing single image based methods which assumed to localize the hand joints in a cropped hand image, we put ourselves in a more practical and complicated scenario. That is, the image we are going to process is a complete image that contains not only multiple hands but also other objects like human body and background. How to build a robust network structure which is adaptive to those variances of images to achieve accurate hand joints detection is still an open problem to be solved. In this paper, we propose a anatomy oriented hand joints detection method with serial-parallel multi-scale feature fusion, which only needs single unmarked image for hand joints detection.

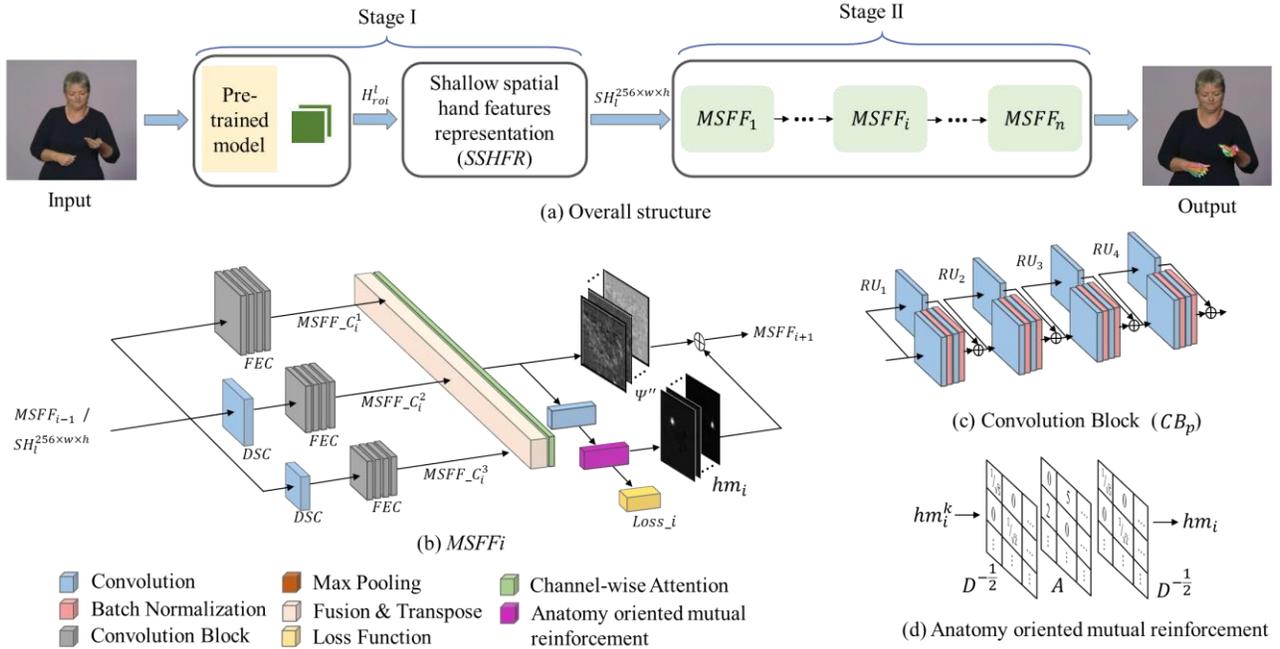

Fig. 1. The architecture of the proposed method. (a) the proposed method is composed of hand localization module, *SSHFR*(Stage I) and parallel-cascade *MSFFs*(Stage II); (b) the structure of a *MSFF* ; (c) the structure of a convolution block ( $CB_p$ ) which consists of four residual units ( $RU_q$ ); (d) Anatomy oriented mutual reinforcement.

## III. PROPOSED METHOD

The proposed method aims to estimate the hand joints position from the input image by designing a robust serial-parallel multi-scale neural network. As shown in Fig.1 (a), this two-stage method locates the joints position $H_l^k(x, y) \in \mathbb{R}^{W \times H}$ from the input $I \in \mathbb{R}^{3 \times W \times H}$, where $l \in [1, L]$ is the number of hands in the image, and $k \in [1, K]$ is the detectable hand joints.

A. *SSHFR (shallow spatial hand features representation)*

The number and size of the hands may be different from image to image. If multi-scale transformation and feature extraction are directly applied to the input image *I*, the network may lose focus and converge slowly. Therefore, the hand regions should be detected and cropped firstly. Firstly, the coarse hand region $H_R^l$ is located by a pre-trained CPM [22]. Then, the precise region of each hand $H_{roi}^l$ is cropped from the input *I*, i.e.,

$$H_{roi}^l = crop(I, H_R^l) \qquad (1)$$

Finally, each $H_{roi}^l$ is fed into a ten-layer convolutional neural network (*SSHFR*) and the local spatial hand features $SH_l$ are generated in *SSHFR*, the convolutional kernels of $Conv_n(n=1,...,5)$ layers are $5\times5$ and those of $Conv_n(n=6,...,10)$ layers are $3\times3$. The output of *SSHFR* is

$$SH_l^{256\times w\times h} = H_{ROI}^l * conv_{1\text{-}10} \qquad (2)$$

where $*$ is the convolution, *w* and *h* are the sizes of output feature maps and *l* means the $l_{th}$ hand in the image. *SSHFR* has 256 output channels. An example of the output of Stage I is shown in Fig.2. Two hand regions $H_{roi}^l\ l\in[1,2]$ are extracted from an image *I* in NZSL dataset and then the shallow feature maps $SH_l^{256\times 64\times 64}$ of each hand are generated through the *SSHFR* structure.

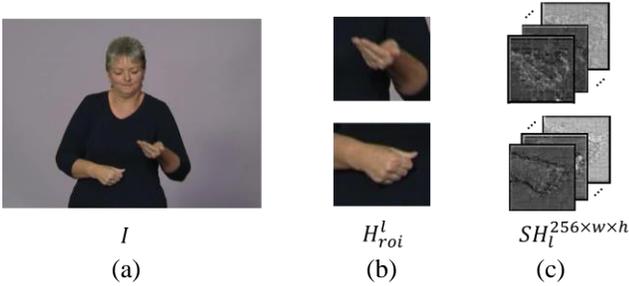

| *I* | $H_{roi}^l$ | $SH_l^{256\times w\times h}$ |
| (a) | (b) | (c) |

Fig. 2. Result of Stage I. (a) An image from NZSL dataset; (b) Detected hand regions by the pre-trained CPM model; (c) Shallow hand features generated by *SSHFR*.

### B. MSFF structure (multiscale feature fusion)

The features at different scales and their fusions are essential for characterizing hand information. The shallow spatial feature of each hand $SH_l^{256\times w\times h}$ is sent to a novel *MSFF* structure. *MSFF* structure generates multi-scale features in parallel and fuses these features, based on the backbone of residual unit.

**MSFF structure** consists of three parallel multi-scale branches $MSFF\_C_i^j$ with similar structure, one features fusion and transformation layer, and one channel-wise attention layer as shown in Fig.1(b), where $i\in[1,...,N]$ is the index of *MSFF* structure, $j\in[1,2,3]$ is multi-scale branch index. Each $MSFF\_C_i^j$ includes a down-sample convolution layer (DSC) and a feature extraction convolution block (FEC). DSC generates different scales feature maps with same network depth through a convolutional kernel size $3\times3$ and stride is 2. Note that the $MSFF\_C_i^1$ only has FEC.

**Feature extraction convolution (FEC)** includes *p* convolutional blocks ($CB_p$) with the same structure, and each $CB_p$ is based on *q* cascaded residual units ($RU_q$), as shown in Fig.1(c). The convolutional kernel size of each $RU_q$ is $3\times3$ and stride is 1. Through modular design, the designed FEC can be reused in different layers of the network. Therefore, the number of $CB_p$ and $RU_q$ can be changed according to the practical task. In this paper, considering the difficulty and efficiency of hand joints detection, we set $p,q\in[1,...,4]$.

**Fusion & Transpose layer:** in each *MSFF* structure, the multi-scale branches generate a series of hand joints feature maps $\psi_i^j$ with different scales. Then, the feature maps generated by $MSFF\_C_i^2$ and $MSFF\_C_i^3$ are up-sampled through bilinear interpolation and concatenated with the feature maps which are produced by $MSFF\_C_i^1$:

$$\Psi_i(c_i\times w_i^1\times h_i^1) = \psi_i^1 + \delta_2(\psi_i^2) + \delta_3(\psi_i^3) \qquad (3)$$

where $w_i^1$ and $h_i^1$ are the feature map sizes of first branch in the $i_{th}$ *MSFF*, and $\delta(\bullet)$ is the up-sampling function.

In order to balance the impact of low-scale features and high-scale features on the hand joints detection accuracy, the channel feature maps of $\Psi_i(c_i\times w_i^1\times h_i^1)$ are reshaped to $\Psi_i(m\times w_i^1\times h_i^1, n\times w_i^1\times h_i^1)$, where $m=3$, $n=c_i/3$. Then, $\Psi_i$ is flatten back to $c_i$ dimension after transpose operation, and concatenated with original $\Psi_i$:

$$\Psi_i^{'} = \xi\left(\Psi_i(m\times w_i^1\times h_i^1, n\times w_i^1\times h_i^1)^T\right) + \Psi_i \qquad (4)$$

where $\xi(\cdot)$ is the flatten operation. Through the transformation of "concatenate-reshape-transpose-reshape-concatenate", the multi-scale channel feature maps which generated by *MSFF* structure are related in context.

The size of the input of *MSFF* is [*b, c, w, h*], where *b* is number of hands which is the output at *SSHFR* stage, *c* is the shallow spatial features of each hand, *w* and *h* are the sizes of feature maps. The feature maps which are produced by $MSFF\_C_i^j (j\in[1,2,3])$ are $c_i^1\times 64\times 64$, $c_i^2\times 32\times 32$ and $c_i^3\times 16\times 16$ respectively. We cascade a series of *MSFF* structure to generate different scale and depth features, and the robustness and detection accuracy of the proposed network are strengthened. The output of $MSFF_i$ is the input of $MSFF_{i+1}$, and number of *MSFF* will be discussed in detail in the section of *Ablation Study*.

**Channel-wise Attention:** The *MSFF* structures produce hand features at different scales. We need to figure out which channels are more meaningful for the next sub-stage. the average-pooling and max-pooling operations are used respectively:

$$\Psi_i^{''} = \Psi_i^{'}\bullet\left(max_{pol}(\Psi_i^{'}) + avg_{pol}(\Psi_i^{'})\right) \qquad (5)$$

Then, $\Psi_i^{''}$ is normalized to $K\times W_i^1\times H_i^1$ by $1\times 1$ convolution, and the heatmap $hm_i^k$ of each hand joint is also generated:

$$hm_i^k = \omega\left(\Psi_i^{*}*conv(K,1)\right) \qquad (6)$$

where $K$ is the number of detectable hand joints, $\omega(\bullet)$ is a normalized function. Through the channel-wise attention mechanism, the hand joint features are reinforced and the training weight of each hand joint heatmap is produced.

**Anatomy oriented mutual reinforcement:** In order to emphasize the joint characteristics in channel feature maps, an anatomy oriented mutual reinforcement mechanism is built, as shown in Fig.1(d). The anatomical relationship of a set of hand joints can be represented by a graph with adjacency matrix $A$ and degree matrix $D$. The heatmap of joint $i$ can be enhanced by the rest heatmaps through their anatomical relationship:

$$hm_i = G(hm_i^1,...,hm_i^k,...,hm_i^{21}, L)$$
$$= hm_i^k + \sum_{k \neq q} l_{k,j} hm_i^q \quad (7)$$

where $L=D^{-1/2}AD^{-1/2}$.

Finally, we calculate the Euclidean distance between the detected hand joint and its ground truth, and make it as the weight of the generated joint heatmap $hm_i^{'k}$ which input to the next MSFF structure.

$$hm_i^{'k} = hm_i^k \cdot \frac{Ed_i^k}{\sum_{k=1}^{K} Ed_i^k} \quad (8)$$

where $Ed_i^k$ is the Euclidean distance between $k_{th}$ hand joint estimated by $MSFF_i$ and the ground truth. The hand joints detection results of the proposed network at Stage II are shown in Fig.3. The $SH_l^{256 \times 64 \times 64}$, which is the output by Stage I, are the inputs of parallel-cascade MSFFs structure. In each MSFF, the normalized hand joint feature maps are produced (Fig.3(a)). Then, the equations (6) and (8) are performed to obtain the heatmap and image coordinates of hand joints (Fig.3(b)). Finally, the network generates the skeleton of each hand as shown in Fig.3(c). We can find that as the number of MSFF structures increases, the proposed network has improved performance of hand joint detection. Some hand joints move towards more accurate positions after going through one or two more MSFF(s), as shown in Fig.3(c).

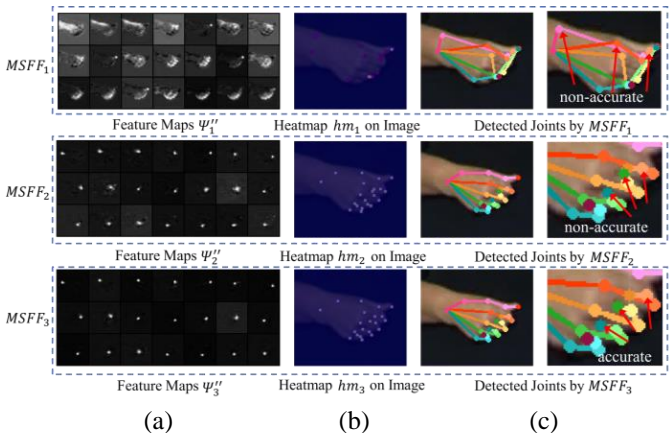

Fig.3 Hand joints location detected by each MSFF (Stage II). (a) the feature maps produced by MSFFs; (b) normalized hand joint heatmaps on hand images; (c) the detected hand joints.

## C. Loss-wise mechanism

During the training, each MSFF has different influence on detection accuracy of hand joints. The loss-wise mechanism can be viewed as producing the weight of each MSFF adaptively. This is important to accelerate the network convergence and improve the hand joint detection accuracy. We map the $hm_i^k$ which output by each MSFF to the image coordinates:

$$MSFF_i(x^k, y^k) = \arg\max_{(x,y) \in \mathbb{R}} hm_i^k(x, y) \quad (9)$$

where $MSFF(x_i^k, y_i^k)$ is the coordinate of the $k_{th}$ hand joint estimated by the $i_{th}$ MSFF structure. Then, the impact of this MSFF on overall network performance is defined as:

$$W_i = \frac{1}{K}\sum_{k=1}^{K} \left\| MSFF_i(x^k, y^k) - G(x^k, y^k) \right\|_2 \quad (10)$$

where $G(x^k, y^k)$ is the ground truth of $k_{th}$ hand joint and $\|\bullet\|_2$ is the L2-norm. The loss of all the MSFFs are combined to form the loss function:

$$Loss = \sum_{i=1}^{N} loss(MSFF_i) * W_i \quad (11)$$

where $N$ is the number of MSFF structure and $loss(MSFF_i)$ is the mean square error loss of each MSFF.

## IV. EXPERIMENTAL RESULTS

### A. Hand Datasets

We evaluate the performance of the proposed network on five datasets: RHD dataset, Hands Synthetic Data (HS), MPII human hand dataset, NZSL dataset and our DCD8-6000 dataset. The numbers of training and testing images in each dataset are listed in TABLE I.

TABLE I
THE NUMBERS OF TRAINING AND TESTING IMAGES IN EACH DATASET.

| Dataset | Type of image | Training images | Testing images | Total |
|---|---|---|---|---|
| RHD | Synthetic | 33,007 | 8,251 | 41,258 |
| HS | Synthetic | 11,409 | 2,852 | 14,261 |
| MPII | Natural | 1,360 | 340 | 1,700 |
| MPII+NZSL* | Natural | 2,560 | 640 | 3,200 |
| DCD8-6000 | Natural | 4,800 | 1,200 | 6,000 |

\* Since the number of images in NZSL is not adequate for conducting the training and testing process, we combine it with MPII as "MPII+NZSL".

The first two datasets, RHD and HS, are synthetic. RHD dataset [9] contains 41,258 images with the resolution of 320×320 PX$^2$. The dataset has RGB images, depth maps, segmentation masks and 19 annotated hand keypoints. In this paper, only the annotated RGB images are used. We randomly select 33,007 training images and reserved the rest 8,251 as testing images. HS dataset [10] has 14,261 hand images with the resolution of 368×368 and 1024×1024 PX$^2$. Each image contains single hand gesture which is annotated by 21 joints. We randomly divide the dataset into a training set (11,409 images) and a testing set (2,852 images).

The MPII and NZSL are natural image datasets. MPII [11], which is collected from YouTube videos, has 1,700 annotated hand images (1280×720 PX$^2$) with various hand gestures of both hands. The New Zealand Sign Language Exercises (NZSL) [12] is a hand gesture dataset used to practice sign language, which contains 1,500 labeled hand joints images with the resolution of 720×576 PX$^2$. The numbers of training and testing images in these two datasets are (1360, 340) and (1200, 300) respectively.

The number of images annotated in MPII and NZSL datasets are limited, while the RHD and Hand Synth datasets are synthetic hand images. Due to illumination, background, etc., the synthetic images and the real-world images have some different characteristics. Moreover, in most of the publicly available human pose detection datasets, the annotation of key points is only up to limb. Therefore, in order to comprehensively verify the performance of the proposed method, we built our own dataset with real hand images and manually annotated the images. Our own dataset is called DCD8-6000, which was collected under a research related to Developmental Coordination Disorder (DCD) [37] [38]. DCD is a type of motor learning difficulty that affects five to six percent of school-aged children, which may have a negative impact on the life of the sufferers. In the experiment, the subject is asked to perform three tasks which are highly related to fine hand movements, i.e. placing pegs, threading lace and drawing trial. The detection of hand joint is essential for improving the efficiency of DCD evaluation. DCD8-6000 dataset includes 6,000 hand images of 8 subjects. The resolution is 960×540 PX$^2$ and the frame rate is 30 f/s. We randomly divide the dataset into a training set (4,800 images) and a testing set (1,200 images). We annotated the 21 hand joints for each hand. Each hand joint is annotated by its coordinate in the image. The occluded joints are annotated to be the position (0, 0). Some sample hand images from DCD8-6000 are shown in Fig.4.

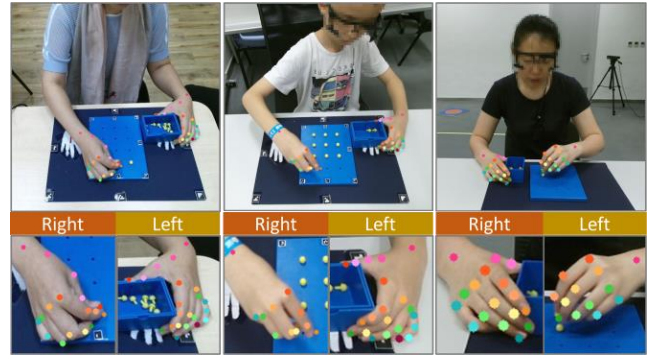

Fig.4. Samples from DCD8-6000 dataset.

### B. Main results

In this section, we present the experimental results in five datasets. The evaluation of our proposed network is carried on an Intel(R) Core(TM) i7-9700 computer with 32 GB RAM and NVIDIA GeForce RTX 2080 GPU.

Some qualitative results of the proposed network on RHD, HS, MPII, NZSL and DCD8-6000 datasets are shown in Fig.5. As we have known, the hand joints detection in the real-world images are more difficult than that in synthetic images. The proposed network has satisfactory detection results in both synthetic and real-world images, as well as in single and multiple hands images. We can also see that the proposed network is robust to self-occlusion (RHD), different hand complexion (HS), complex hand gestures (DCD8-6000 and NZSL), and the background changes (MPII).

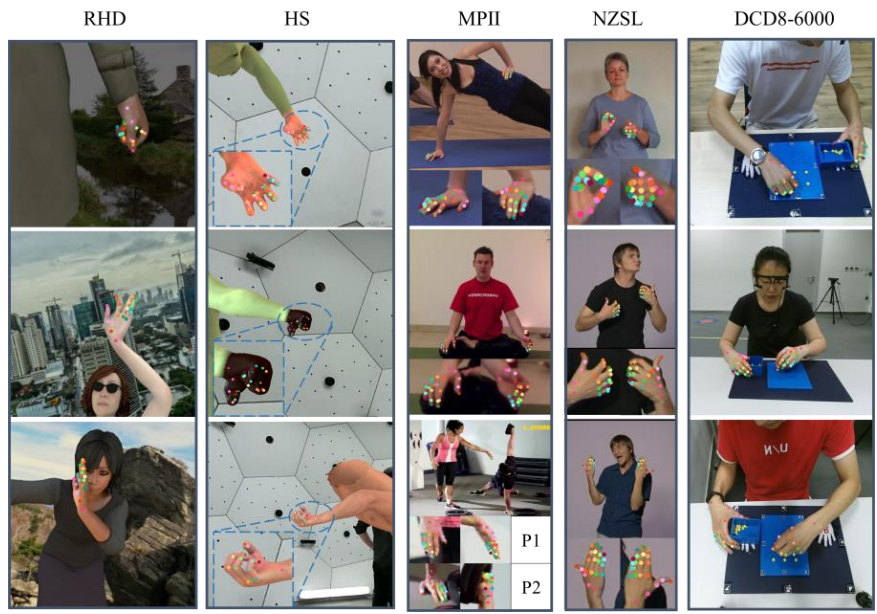

Fig.5. Some hand joints detection results. The first two columns are synthetic images from RHD and HS datasets, and the last three columns are real-world hand images from MPII, NZSL and DCD8-6000 datasets (*P1, P2 represents the first and second person, respectively).

The Probability of Correct Keypoints (PCK) is used to measure the detection accuracy of the hand joints. PCK is the probability of the estimated hand joint within a normalized distance $\tau$ to its ground truth, defined as

$$PCK_\tau^k = \frac{1}{T}\sum_T \sigma\left(\sqrt{\left((x_i^k - g_{xi}^k) + (y_i^k - g_{yi}^k)\right)^2} < \tau\right) \quad (11)$$

where $T$ is the size of validation dataset and $\sigma$ is a normalization function. In [10] and [20], the head size was used as the normalization factor. However, the head sizes information is unavailable in RHD, HS and DCD8-6000 datasets (only hands in images). Therefore, we use the size of hand bounding box of the ground truth as the normalization factor. Fig.6 presents the average hand joints detection rates under different normalization distances $\tau \in [0:0.1:0.5]$, by our proposed network on the five datasets.

The detection rates of hand joints on synthetic dataset RHD and HS are 94% and 92% respectively, while on MPII and MPII+NZSL are 71% and 84% at PCK@0.2. This may be because RHD and HS have more high quality training samples than MPII and NZSL. The performance on DCD8-6000 is better than other datasets. In the DCD8-6000 dataset, the output resolution of hand regions in Stage I (150×150 pixels) is higher than others (that of MPII and NZSL is 90×90 pixels), which provides more detailed features for the *MSFFs* in Stage II. Moreover, the subjects performed same tasks in DCD8-6000 dataset, so the consistency between training samples and testing samples is better than those of other datasets. Overall speaking, the average hand joints detection rate is greater than 90% at PCK@0.3 in all datasets, and that means the performance of the proposed network is satisfactory.

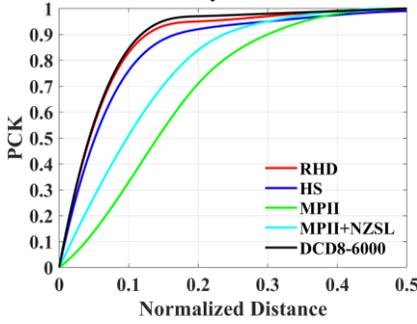

Fig. 6. PCK curves on the RHD, HS, MPII, MPII+NZSL and DCD8-6000 datasets.

TABLE II
COMPARISON WITH STATE-OF-THE-ART ALGORITHMS ON RHD DATASET.

| Method | PCK @0.1 | PCK @0.2 | PCK @0.3 | PCK @0.4 | PCK @0.5 |
|---|---|---|---|---|---|
| CPM[22] | 0.67 | 0.78 | 0.86 | 0.94 | 0.96 |
| OPH[25] | 0.12 | 0.23 | 0.34 | 0.51 | 0.82 |
| HRnet[36] | 0.83 | 0.92 | 0.98 | 0.98 | 0.99 |
| Ours | **0.84** | **0.94** | **0.98** | **0.99** | **0.99** |

TABLE III
COMPARISON WITH STATE-OF-THE-ART ALGORITHMS ON HS DATASET.

| Method | PCK @0.1 | PCK @0.2 | PCK @0.3 | PCK @0.4 | PCK @0.5 |
|---|---|---|---|---|---|
| CPM[22] | 0.53 | 0.86 | 0.97 | 0.99 | 1 |
| OPH[25] | 0.34 | 0.59 | 0.77 | 0.93 | 0.99 |
| HRnet[36] | 0.72 | 0.89 | 0.95 | 0.99 | 1 |
| Ours | **0.76** | **0.92** | **0.97** | **1** | **1** |

TABLE IV
COMPARISON WITH STATE-OF-THE-ART ALGORITHMS ON MPII+NZSL DATASET.

| Method | PCK @0.1 | PCK @0.2 | PCK @0.3 | PCK @0.4 | PCK @0.5 |
|---|---|---|---|---|---|
| CPM[22] | 0.38 | 0.61 | 0.81 | 0.96 | 0.98 |
| OPH[25] | 0.31 | 0.57 | 0.78 | 0.81 | 0.86 |
| HRnet[36] | 0.47 | 0.81 | 0.93 | 0.97 | 0.99 |
| Ours | **0.51** | **0.84** | **0.95** | **0.99** | **1** |

TABLE V
COMPARISON WITH STATE-OF-THE-ART ALGORITHMS ON DCD8-6000 DATASET.

| Method | PCK @0.1 | PCK @0.2 | PCK @0.3 | PCK @0.4 | PCK @0.5 |
|---|---|---|---|---|---|
| CPM[22] | 0.38 | 0.49 | 0.80 | 0.97 | 0.98 |
| OPH[25] | 0.27 | 0.35 | 0.76 | 0.85 | 0.89 |
| HRnet [36] | 0.82 | 0.92 | 0.95 | 0.96 | 0.98 |
| Ours | **0.88** | **0.97** | **0.98** | **0.99** | **1** |

To better demonstrate the performance of the proposed network, we compare the hand joints detection rates with the state-of-the-art networks, i.e. CPM [22], OpenPose-hand (OPH) [25] and HRnet[36]. The input of CPM, OPH and HRnet is supposed to be a pre-detected hand box. To make the comparison fair, we crop hands regions from the images, resize them to 256×256 pixels, and pre-train their networks. The comparison results are listed in TABLEs II-V. Our network achieves 95% PCK@0.3 or above on all the datasets, which over performs other methods.

In order to verify the efficiency of our proposed method, the running time of the proposed network is shown in TABLE VI in detail. The processing time of Stage I depends on the resolution of the input images. The processing time of Stage I is 17ms for RHD (resolution 320×320 pixels), and that is 31ms for MPII (resolution 1280×720 pixels). The processing time of parallel-cascaded *MSFF* structure depends on the number of hands in the image. The average processing time of $MSFF_{1-3}$ structures for one hand is around 10.4ms. Overall speaking, assuming one hand in the image, the number of frames per second the method can process is between 24ms and 37ms. This is adequate for supporting many real-time applications. More detailed results are shown in the following TABLE VI.

TABLE VI
THE PROCESSING TIME OF PROPOSED NETWORK IN DIFFERENT DATASETS.

| Dataset | Image resolution ($PX^2$) | Stage I (ms) | $MSFF_{1-3}$ (ms) | Frames per Second |
|---|---|---|---|---|
| RHD | 320×320 | 17 | 10 | 37 |
| HS | 368×368 | 19 | 10 | 34 |
| MPII | 1280×720 | 31 | 11 | 24 |
| NZSL | 720×576 | 25 | 11 | 28 |
| DCD8-6000 | 960×540 | 28 | 10 | 26 |

| | Average | 24 | 10.4 | 29.8 |
|---|---|---|---|---|

## C. Ablation Study

The ablation study is conducted on MPII+NZSL dataset to investigate the contributions of the function modules of the proposed network, and the results are presented in Fig.7. The details of the ablation are as followings. *Del SSHFR*: SSHFR is skipped and image batches are directly input to *MSFF* structures. *Del TP*: the transpose layer of multi-scale features is removed in each *MSFF*. *Del AT*: the channel-wise attention layers are removed from *MSFFs* structure. *Del LW*: loss-wise mechanism of each *MSFF* during the training is removed, and only the last *MSFF* is used to calculate loss function, as in HRnet [36]. *Del AOMR*: the anatomy oriented mutual reinforcement is removed from *MSFFs* structure.

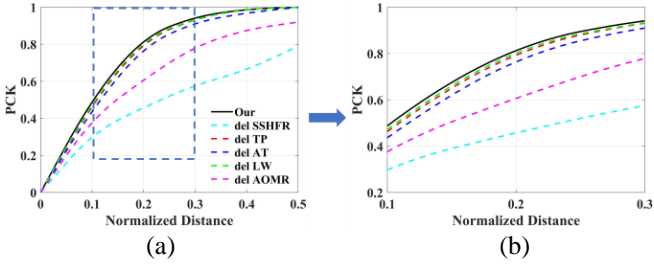

Fig.7. Ablation study. (a) the influence of each module on hand joints detection rate at different PCKs (3 *MSFFs* and 60 training epochs); (b) the enlarged blue dotted region of (a).

All the modules contribute to the performance of the proposed network, in which *SSHFR* has the greatest impact on network performance. This is because if we use parallel-cascade *MSFF* structure directly, the shallow multi-scale network is difficult to produce deep hand features effectively. The spatial correlation between different features are established through *TP* operation, which helps to improve the performance. In addition to *SSHFR*, the joint spatial connection makes the network focus on those inaccurate joints during the training. Adding *AOMR* changes PCK@0.1 from 39% to 51%. The loss-wise mechanism generates the training weight of each *MSFF* adaptively, which also can improve the hand joints detection accuracy.

## D. Number of MSFFs

Theoretically, the network can be extended indefinitely by adding *MSFF*s. We study the relationship between the network performance and the number of *MSFFs* on MPII+NZSL dataset. Fig.8(a-e) show the PCK curves at normalize distance $\tau \in [0:0.5]$ over different number of *MSFFs*. We can find that the detection rate increases monotonically until three *MSFFs*. However, if the number of *MSFF* continues to increase, the growth of performance tends to be saturated, which may be caused by overfitting or local optimization. If the number of *MSFFs* increases, the network training and testing time also increase. To balance the performance and computational cost, three *MSFFs* are used for MPII+NZSL dataset.

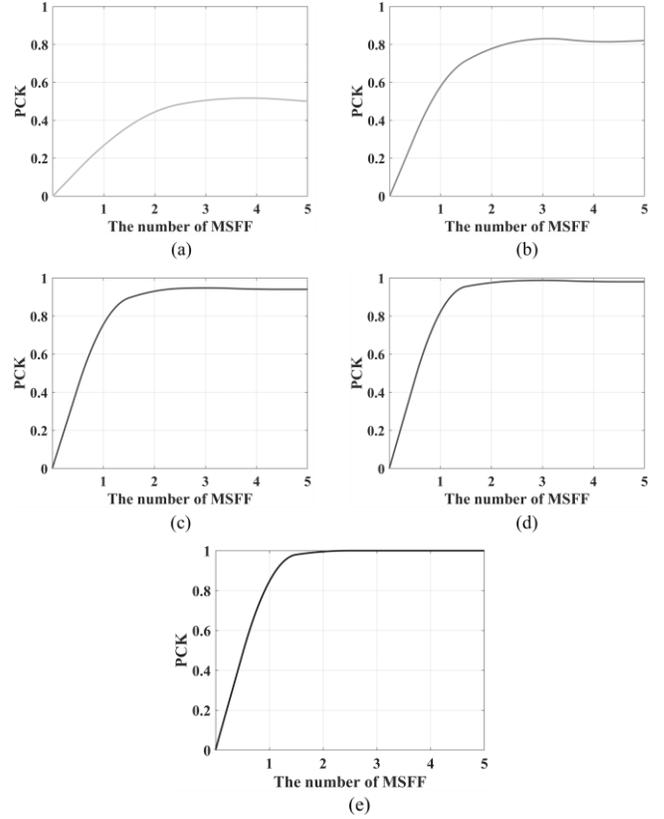

Fig.8. Comparisons of the number of *MSFFs* on MPII+NZSL dataset. (a) PCK@0.1; (b) PCK@0.2; (c) PCK@0.3; (d) PCK@0.4; (e) PCK@0.5 with different number of *MSFFs*.

## E. Confidence of detected hand joint

We observed an interesting phenomenon in the experiment: the shape of the heatmap seems to be related to detection rate. As shown in Fig. 3., the white parts of heatmaps after $MSFF_1$ occupy large portion of the whole map. At the same time, the detection result is not accurate. After going through more *MSFFs*, the white parts concentrate on a smaller area on the map, and the detection result becomes more accurate. This observation inspires us to define a statistic of the heatmap, called "spread" to indicate the accuracy of the detection result, i.e.,

$$spread_k = nor\left(\sum\nolimits_{(x,y)\in \mathbb{R}} \left\|(hm_i^k(x,y)>0) - MSFF_i(x^k, y^k)\right\|_2\right)$$
(12)

where $nor(\bullet)$ is a normalization function. Spread is actually the average distance of the non-zero points to the detected joint. We study the relationship between the spread of the heatmap and detection accuracy of each hand joint in MPII+NZSL dataset, as shown in Fig. 9. The larger the spread of the heatmap, the lower is the detection accuracy, which means spread can be used as an indicator of the detection confidence.

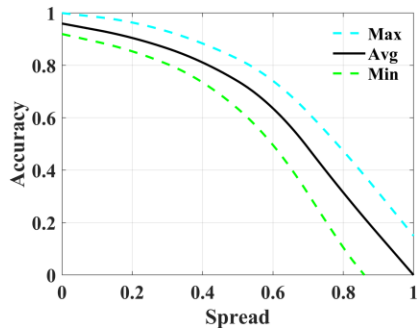

Fig.9. The relationship between detection accuracy and spread of the heatmap. The black solid line is the average accuracy. The cyan and green dotted line are the maximum and minimum accuracies respectively.

## V. Conclusion

In this paper, a novel anatomy-oriented hand joints detection with cascaded multiscale feature fusion network is proposed. The innovations of our proposed network are: (1) it is an single image-based end-to-end hand joints detection network, which does not need any prior knowledge; (2) the shallow features and hand location features extracted in Stage I are fully utilized by *MSFFs* structure, which improves the accuracy; (3) the combination of parallel-cascade multiscale transformation and channel-wise attention mechanism improves the efficiency of features selection; (4) the outputs of each *MSFF* contribute to the loss function; (5) the heatmaps are enhanced with each other by adopting anatomical relationship. The improvements brought by the new proposed structures have been proved by comprehensive experiments on five datasets. The performance of our network is competitive to the state-of-the-arts. It also provides new solutions to similar computer vision problems, such as human pose detection and hand gesture recognition.